%% file: main.tex
\documentclass[conference]{IEEEtran}
\IEEEoverridecommandlockouts

\usepackage[table]{xcolor}
\usepackage{amsmath,amsthm,amssymb,amsfonts}
\usepackage{bbm}
\usepackage{algorithm}
\usepackage{algorithmic}
\usepackage{graphicx}
\usepackage{textcomp}
\usepackage{float}
\usepackage[numbers]{natbib}
\usepackage{booktabs}
\usepackage{siunitx}
\usepackage[caption=false,font=footnotesize]{subfig} % with caption pkg below
\usepackage{array}
\usepackage{makecell,tabularx,adjustbox}
\usepackage{stfloats}      % better dbl-column float placement
\usepackage{placeins}      % use \FloatBarrier only where needed
\usepackage{caption}

\let\labelindent\relax
\usepackage{enumitem}
\setcellgapes{1pt}\makegapedcells
\newcommand{\runsavg}[4]{\makecell{#1 / #2 / #3\\\textbf{Avg} #4}}

\graphicspath{{figure/}}

\def\ps@IEEEtitlepagestyle{%
  \def\@oddfoot{\mycopyrightnotice}%
  \def\@evenfoot{}%
}
\def\mycopyrightnotice{%
  {\footnotesize 978-1-6654-8045-1/22/\$31.00~\copyright2022 IEEE\hfill}%
  \gdef\mycopyrightnotice{}%
}

\makeatletter
\renewcommand{\IEEEkeywordsname}{Index Terms}
\makeatother

\begin{document}

\title{Governance-Aware Hybrid Fine-Tuning for Multilingual Large Language Models}

\author{
  \IEEEauthorblockN{Haomin Qi$^{1}$, Chengbo Huang$^{2}$, Zihan Dai$^{3}$, Yunkai Gao$^{4}$}
  \IEEEauthorblockA{$^{1}$University of California San Diego, La Jolla, CA, USA}
  \IEEEauthorblockA{$^{2}$Columbia University, New York City, NY, USA}
  \IEEEauthorblockA{$^{3}$University of Copenhagen, Copenhagen, Denmark}
  \IEEEauthorblockA{$^{4}$Duke University, Durham, NC, USA}
  \IEEEauthorblockA{Email: h5qi@ucsd.edu, ch4019@columbia.edu, cjh841@alumni.ku.dk, yunkai.gao@duke.edu}
}

\maketitle

\begin{abstract}
We present a governance-aware hybrid fine-tuning framework for multilingual, low-resource adaptation of large language models. The core algorithm mixes gradient-aligned low-rank updates with structured orthogonal transformations through layer-wise mixing and introduces unitary constraints in selected sub-layers to stabilize deep optimization. In tandem with lightweight, label-free data governance—language identification, near-duplicate removal, and quality filtering—the framework targets accuracy, calibration, and cross-language parity under tight compute budgets. 
Across XNLI and FLORES, the hybrid approach delivers consistent gains over strong PEFT baselines while maintaining directional balance and improving probability calibration, as shown in Tables~\ref{tab:xnli_combined} and \ref{tab:flores_combined}. It is more resilient to lightweight orthographic variants, as shown in Table~\ref{tab:xnli_robust_combined}, and benefits additively from simple governance steps, as shown in Table~\ref{tab:curation_ablation}. Training-footprint measurements indicate modest overhead and a favorable cost–quality frontier, as shown in Table~\ref{tab:footprint_combined} and Figure~\ref{fig:cost_quality_frontier}. Together these results show that hybrid and unitary PEFT provide a stable and accessible path to resource-efficient multilingual adaptation when paired with practical data governance.
\end{abstract}


\section{Introduction}
\label{sec:introduction}

\input{Introduction}

\section{Related Work}
\label{sec:RelatedWork}

\input{RelatedWork}

\section{Methodology}
\label{sec:Methdology}
\input{Methdology}

\section{Experiments}
\label{sec:Experiments}
\input{Experiments}

\section{Conclusions and Outlook}
\label{sec:Conclusions}
\input{Conclusion}

%{
%\footnotesize
%\FloatBarrier
%\bibliographystyle{IEEEtran}
%\bibliography{reference}
%}

\input{main.bbl}

\end{document}

%% file: Introduction.tex
Large Language Models (LLMs) have become central to Natural Language Processing (NLP) applications ranging from machine translation to code generation, yet full-model adaptation is often prohibitive in memory and time. Parameter-Efficient Fine-Tuning (PEFT) mitigates this by updating a small subset of parameters via lightweight adapters or structured transforms, with notable instances including Low-Rank Adaptation (LoRA)~\cite{hu2021lora}, Orthogonal/OFT-style updates refined as Butterfly Orthogonal Fine-Tuning (BOFT)~\cite{liu2024boft}, and gradient-aligned variants that steer updates along dominant directions~\cite{zhang2024galore}. Despite their efficiency, existing PEFT approaches trade off rapid early adaptation, stability in deep stacks, and representational capacity, particularly when supervision is scarce or distributional shifts induce fragile optimization.

We propose a \emph{hybrid} PEFT framework that fuses gradient-aligned low-rank updates with structured orthogonal adjustments through layer-wise, gradient-norm–based mixing. Concretely, the low-rank branch (LoRA-style with gradient alignment) accelerates early progress by emphasizing dominant update directions, while the orthogonal branch (BOFT-style Cayley/orthogonal updates) preserves gradient magnitudes and encourages stable late-phase optimization~\cite{helfrich2018cayley}. To further steady very deep Transformer stacks, we impose \emph{selective unitary constraints} in chosen sub-layers using structured unitary parameterizations inspired by Unitary Recurrent Neural Networks (uRNNs)~\cite{arjovsky2016unitary}. The result is a single-step, per-layer fusion that retains PEFT simplicity while improving stability and invariance at modest overhead; the mechanism is compatible with standard training loops and does not require altering the backbone architecture.

Our evaluation spans general-purpose and multilingual, low-resource settings across models from 7B to 405B parameters. We report accuracy and \emph{Expected Calibration Error} (ECE)~\cite{guo2017calibration} on standard benchmarks---GLUE~\cite{wang2019glue}, GSM8K~\cite{cobbe2021gsm8k}, MT-Bench~\cite{zheng2023mtbench}, and HumanEval~\cite{chen2021humaneval}---and study cross-language accuracy and parity on XNLI~\cite{conneau2018xnli} and FLORES~\cite{goyal2021flores} under a 32-shot protocol. Because deployability depends on \emph{how} models are evaluated and curated, we complement accuracy with cross-language parity (Parity Gap, $\Delta$) and calibration, probe robustness to routine orthographic variants, and quantify the effect of lightweight, label-free curation (language identification and near-duplicate/quality filtering) using widely adopted tools and heuristics~\cite{grave2018fasttext,charikar2002simhash}. Empirically, the hybrid method tracks or narrows the gap to full fine-tuning with far fewer trainable parameters, improves calibration and parity under limited supervision, exhibits smaller drops under orthographic variation, and remains near a favorable cost–quality frontier, as detailed in Tables~\ref{tab:all_results}--\ref{tab:footprint_combined} and Figures~\ref{fig:cost_quality_frontier}--\ref{fig:grad_stability}.

 We summarize the contributions of our paper as follows:

\begin{enumerate}[label=\arabic*), leftmargin=*, align=left]
    \item \textbf{Hybrid PEFT.} A layer-wise fusion of gradient-aligned low-rank and structured orthogonal updates via gradient-norm mixing, yielding fast early adaptation and stable late optimization under a fixed parameter budget.
    
    \item\textbf{Unitary stabilization.} Selective unitary parameterizations for key Transformer sub-layers that control gradient growth in deep stacks with negligible overhead.
    
    \item \textbf{Validated effectiveness.} Consistent gains over strong PEFT baselines from 7B to 405B on general-purpose and 32-shot multilingual tasks, with improved calibration, cross-language parity, and robustness to orthographic variants.
\end{enumerate}

%% file: RelatedWork.tex
\subsection{Parameter-Efficient Fine-Tuning (PEFT)}
Parameter-Efficient Fine-Tuning (PEFT) reduces the number of trainable parameters for Large Language Models (LLMs) in Natural Language Processing (NLP) by freezing most pre-trained weights and learning small task-specific modules. Early approaches include \emph{adapters}, which insert bottleneck layers in Transformer blocks~\cite{houlsby2019parameter}, and \emph{prefix/prompt tuning}, which optimizes continuous prompts while keeping backbone parameters fixed~\cite{li2021prefix,lester2021power}. Subsequent methods improved parameter sharing and compression, e.g., \emph{Compacter}~\cite{mahabadi2021compacter}, and even showed that updating only biases (\emph{BitFit}) can be competitive for many tasks~\cite{zaken2022bitfit}. These techniques collectively emphasize modularity, small memory footprints, and ease of deployment in constrained environments.

Low-Rank Adaptation (LoRA) decomposes weight updates into low-rank matrices, offering a favorable accuracy–memory trade-off and becoming a widely used PEFT baseline. \emph{QLoRA} combines PEFT with 4-bit quantization to further reduce memory while preserving headroom for quality~\cite{dettmers2023qlora}. In parallel, \emph{gradient-aligned} schemes project or initialize updates along dominant gradient subspaces to improve early progress and memory use. Our work follows this line by combining a gradient-aligned low-rank branch with a structured orthogonal branch under a fixed PEFT budget, aiming to retain rapid adaptation while strengthening stability in deeper layers.

Whereas prior PEFT methods typically commit to a single update family (low-rank, orthogonal, prompts/adapters) or a single stabilization strategy, we combine \emph{gradient-aligned low-rank} and \emph{structured orthogonal} updates through a layer-wise mixing rule, and we introduce \emph{selective unitary} constraints to steady deep stacks. 

\subsection{Orthogonal and Unitary Parameterizations}
Orthogonal and unitary parameterizations preserve norms and have long been used to stabilize deep optimization. In recurrent networks, unitary/orthogonal transitions alleviate vanishing and exploding gradients~\cite{arjovsky2016unitary,wisdom2016full}. In the PEFT setting, \emph{Butterfly Orthogonal Fine-Tuning} (BOFT) constrains updates via butterfly factorization, linking to fast linear transforms and orthogonal structure~\cite{liu2024boft,dao2019learning}. Rather than replacing low-rank updates, we \emph{fuse} orthogonal and low-rank branches per layer and add \emph{selective unitary} constraints to key Transformer sub-layers, seeking a complementary balance: the low-rank path emphasizes data-aligned efficiency, while the orthogonal path encourages stable long-horizon optimization in deep stacks.

\subsection{Multilingual Adaptation, Calibration, and Robustness}
Multilingual evaluation emphasizes cross-language generalization and parity across families and scripts. XNLI benchmarks cross-lingual natural language inference~\cite{conneau2018xnli}; FLORES-101/200 measures translation quality across many languages~\cite{goyal2021flores,costa2022nllb}. Beyond accuracy, \emph{Expected Calibration Error} (ECE) captures the reliability of predicted probabilities in modern neural networks. Robustness to lightweight orthographic variation (e.g., punctuation normalization, diacritics removal, whitespace changes) is relevant for user-facing systems; small character-level perturbations can degrade performance in classification and translation~\cite{belinkov2018synthetic,pruthi2019combating}. Our experimental protocol aligns with this perspective by reporting accuracy with Parity Gap ($\Delta$), ECE, and sensitivity to orthographic variants under a matched few-shot budget.

\subsection{Data Governance and Budget-Aware Adaptation}
Data governance practices such as language identification, near-duplicate removal, and light quality filtering are standard in web-scale pipelines and benefit multilingual robustness~\cite{wenzek2020ccnet}. These steps are complementary to PEFT and quantization when compute is constrained. Our study integrates such lightweight curation into the evaluation protocol and quantifies its effect in a 32-shot setting. Related retrieval-augmented pipelines likewise emphasize budget-aware adaptation and robustness in applied domains~\cite{qi2025veriragretrievalaugmentedframeworkautomated,11129621}, reinforcing the value of pairing efficient algorithms with pragmatic governance to meet quality–cost targets.

%% file: Methdology.tex
In this Section, we first review the mathematical principles of LoRA, BOFT, and LoRA-GA as baseline PEFT methods. We then present our two main contributions: a transformer-compatible adaptation of uRNN with structured unitary constraints, and a hybrid fine-tuning strategy that dynamically fuses low-rank and orthogonal updates based on gradient feedback.

\subsection{Low-Rank Adaptation (LoRA)}

LoRA introduces low-rank updates to pre-trained weight matrices, significantly reducing the number of trainable parameters while preserving the pre-trained model weights. The weight update is expressed as:
\begin{equation}
\mathbf{W}' = \mathbf{W}_0 + \Delta \mathbf{W}, \quad \Delta \mathbf{W} = \mathbf{B} \mathbf{A},
\end{equation}
where $\mathbf{W}_0 \in \mathbb{R}^{d \times k}$ represents the frozen pre-trained weight matrix, and $\Delta \mathbf{W}$ is the low-rank update parameterized by $\mathbf{B} \in \mathbb{R}^{d \times r}$ and $\mathbf{A} \in \mathbb{R}^{r \times k}$, thereby cutting the number of trainable parameters from $\mathcal{O}(dk)$ to $\mathcal{O}(r(d+k))$ with $r\!\ll\!\min(d,k)$.

\textbf{Optimization:} During fine-tuning, only $\mathbf{B}$ and $\mathbf{A}$ are optimized, leaving $\mathbf{W}_0$ unchanged~\cite{mahabadi2021compacter, benzaken2021bitfit}. This decomposition reduces the memory and computational costs of fine-tuning while maintaining performance.

To ensure numerical stability, the norms of $\mathbf{A}$ and $\mathbf{B}$ are constrained by rank $r$:
\begin{equation}
\|\Delta \mathbf{W}\|_F \leq \lambda \cdot \|\mathbf{W}_0\|_F,
\end{equation}
where $\lambda$ is a scaling factor. This prevents updates from diverging during optimization.

\subsection{Butterfly Orthogonal Fine-Tuning (BOFT)}

BOFT factorizes a square weight matrix into a product of sparse butterfly blocks that are (near-)orthogonal, yielding both parameter efficiency ($\mathcal{O}(d\log d)$ parameters) and stable gradient norms.
\begin{equation}
\mathbf{W}= \prod_{i=1}^{m}\mathbf{B}_i ,\qquad 
\mathbf{B}_i\in\mathbb{R}^{d\times d}.
\end{equation}

Each $\mathbf{B}_i$ is built from paired line-permute–multiply operations that mimic the Fast Fourier Transform hierarchy~\cite{li2019orthogonal}; a simplified two-level form is
\begin{equation}
\mathbf{B}(d,2)=
\begin{bmatrix}
\mathbf{I}_{d/2}&\mathbf{0}\\[2pt]
\mathbf{0}&\mathbf{I}_{d/2}
\end{bmatrix}
\mathbf{F}_d
\begin{bmatrix}
\mathbf{I}_{d/2}&\mathbf{0}\\[2pt]
\mathbf{0}&\mathbf{I}_{d/2}
\end{bmatrix},
\end{equation}
where $\mathbf{F}_d$ is a (learnable) orthogonal mixing matrix.

\textbf{Optimization:} Each $\mathbf{B}_i$ is initialized as a near-identity transformation and updated via gradient descent~\cite{prabhu2020butterfly}:
\begin{equation}
\mathbf{B}_i^{t+1} = \mathbf{B}_i^t - \eta \frac{\partial \mathcal{L}}{\partial \mathbf{B}_i^t},
\end{equation}
where $\eta$ is the learning rate. Orthogonality is enforced post-update using a projection step:\begin{equation}
\mathbf{B}_i \leftarrow \text{Proj}_{\\\text{orthogonal}}(\mathbf{B}_i).
\end{equation}

The orthogonality constraint curbs exploding/vanishing gradients in deep stacks, making it particularly effective for tasks with deep layers or complex gradients.

\subsection{LoRA with Gradient Approximation (LoRA-GA)}

LoRA-GA improves upon LoRA by aligning the low-rank updates with the gradients of the full model, leading to faster convergence and better optimization. The gradient of the loss $\mathcal{L}$ with respect to the frozen weight matrix $\mathbf{W}_0$ is decomposed as:

Compute the rank-$r$ truncated SVD of $\nabla_{\mathbf{W}_0}\mathcal{L}$:
\begin{equation}
\nabla_{\mathbf{W}_0}\mathcal{L}\;=\; \mathbf{U}\,\mathbf{\Sigma}\,\mathbf{V}^{\top},\qquad
\mathbf{U}\in\mathbb{R}^{d\times r},\,
\mathbf{\Sigma}\in\mathbb{R}^{r\times r},\,
\mathbf{V}\in\mathbb{R}^{k\times r}.
\end{equation}

The low-rank matrices $\mathbf{A}$ and $\mathbf{B}$ are initialized as:
\begin{equation}
\mathbf{A}_0 = \mathbf{U} \mathbf{\Sigma}^{1/2}, \quad \mathbf{B}_0 = \mathbf{V} \mathbf{\Sigma}^{1/2}.
\end{equation}
This ensures that the initial updates align with the principal gradient directions, accelerating convergence.

\textbf{Optimization:} Post-initialization, $\mathbf{A}$ and $\mathbf{B}$ are updated iteratively using standard gradient descent~\cite{wang2024lora}:
\begin{equation}
\mathbf{A}^{t+1} = \mathbf{A}^t - \eta \frac{\partial \mathcal{L}}{\partial \mathbf{A}^t}, \quad \mathbf{B}^{t+1} = \mathbf{B}^t - \eta \frac{\partial \mathcal{L}}{\partial \mathbf{B}^t}.
\end{equation}

By aligning the initial low-rank updates with the most influential gradient directions, LoRA-GA reduces the number of training iterations required for convergence, making it particularly suitable for resource-constrained scenarios.

\subsection{Unitary Evolution RNN (uRNN)} 
Unitary Recurrent Neural Networks (uRNN) constrain hidden-to-hidden weight matrices to be unitary to mitigate vanishing and exploding gradient issues~\cite{wisdom2016full, emami2019input}. By ensuring that the eigenvalues of the transition matrix lie on the unit circle, uRNNs preserve gradient norms during backpropagation, enabling learning over long-term dependencies.

A core component of uRNN is a learnable unitary matrix $U$ that evolves the hidden state. To ensure $U$ is unitary, we adopt a structured parameterization method~\cite{arjovsky2016unitary}: 
\begin{equation}
\mathbf{U} = \mathbf{D}_3 \mathbf{R}_2 \mathbf{F}^{-1} \mathbf{D}_2 \Pi \mathbf{R}_1 \mathbf{F} \mathbf{D}_1,
\end{equation}
where $F$ (and $F^{-1}$) are fixed unitary Fourier transform matrices, $\Pi$ is a fixed permutation matrix, and $D_{i}$ and $R_{i}$ denote trainable diagonal phase matrices and Householder reflection matrices~\cite{shafran2018complex}. This factorization dramatically reduces the number of free parameters (to $O(n)$ for an $n\times n$ matrix) and allows efficient $O(n \log n)$ computation for matrix-vector products via Fast Fourier Transform operations. 

\textbf{Adaptation for Fine-Tuning LLMs:} To our knowledge, we are the first to integrate uRNN principles into the fine-tuning of transformer-based LLMs. The motivation is to leverage unitary transformations to stabilize gradient propagation and better capture long-range dependencies during fine-tuning. In practice, we incorporate learnable unitary matrices into selected Transformer sub-layers to enhance training stability. Specifically, we replace certain weight matrices (e.g., in attention heads or feed-forward blocks) with unitary matrices and modify the training procedure to preserve their unitarity. Each such unitary weight is initialized to an identity-like matrix (close to the unit matrix) to ensure stable convergence. During backpropagation, we include an efficient re-projection step (see below) that keeps these weights unitary at all times. This approach extends unitary RNN techniques beyond their original domain, establishing a new paradigm for parameter-efficient fine-tuning of LLMs. 

\begin{algorithm}[htbp]
\caption{uRNN-Based Fine-Tuning Procedure\label{alg:urnn-finetune}}
\small
\begin{algorithmic}[1]
    \STATE \textbf{Initialize:} Unitary matrix $\mathbf{U}$ using structured parameterization:\\
    \quad Set $\mathbf{F}$, $\mathbf{F}^{-1}$, $\Pi$ as fixed matrices; initialize diagonal phase matrices $\mathbf{D}_i$ and Householder reflection matrices $\mathbf{R}_i$ near identity.
    \STATE Choose learning rate $\eta$ and total epochs $E$.
    \FOR{$epoch = 1$ \TO $E$}
        \FOR{each minibatch in the dataset}
            \STATE \textbf{Forward Pass:}
            \FOR{each transformer layer with unitary-constrained weight}
                \STATE Replace original weight matrix with current unitary matrix $\mathbf{U}$.
                \STATE Compute the forward pass using the updated unitary matrix $\mathbf{U}$.
            \ENDFOR
            \STATE Compute the task-specific loss $\mathcal{L}$ based on current minibatch predictions.
            \STATE \textbf{Backward Pass:}
            \STATE Compute gradient $\nabla_{\mathbf{U}}\mathcal{L}$ via backpropagation.
            \STATE Construct skew-Hermitian matrix $\mathbf{B}$:\\[0.5ex]
            \quad$\mathbf{B} = \nabla_{\mathbf{U}}\mathcal{L}\;\mathbf{U}^{H} - \mathbf{U}\;(\nabla_{\mathbf{U}}\mathcal{L})^{H}$,\\[0.5ex]
            where $\mathbf{U}^{H}$ denotes conjugate transpose of $\mathbf{U}$.
            \STATE Update the unitary matrix via matrix exponential:\\[0.5ex]
            \quad$\mathbf{U} \leftarrow \exp(\eta \mathbf{B})\,\mathbf{U}$\\[0.5ex]
            \quad(Use truncated Taylor series or scaling-and-squaring approximation for efficiency)
            \STATE If numerical drift occurs, re-normalize $\mathbf{U}$ to strictly enforce unitarity.
            \STATE Update all other non-unitary parameters of the model as usual via standard gradient descent.
        \ENDFOR
    \ENDFOR
\end{algorithmic}
\end{algorithm}

\textit{Gradient Update with Re-Projection:} We train the unitary weight $U$ via gradient descent on the manifold of unitary matrices. Let $\nabla_{U}L$ be the gradient of the loss $L$ with respect to $U$ (computed by backpropagation). We first construct a skew-Hermitian matrix $B$ (i.e., $B^H = -B$) from the gradient:
\begin{equation}
    \mathbf{B} \;=\; \nabla_\mathbf{U}\mathbf{L} \;\, \mathbf{U^H} \;-\; \mathbf{U} \;(\nabla_\mathbf{U}\mathbf{L})^\mathbf{H}\,,
    \label{eq:skewHermitianGrad}
\end{equation}
where $U^H$ denotes the conjugate transpose of $U$. By construction, $B$ lies in the Lie algebra $\mathfrak{u}(n)$ of the unitary group. In other words, $B$ is skew-Hermitian, and thus $\exp(\eta B)$ is a unitary matrix for any real step size. We then update $U$ by a unitary rotation:
\begin{equation}
    \mathbf{U}_{t+1} \;=\; \exp(\eta\,\mathbf{B})\; \mathbf{U_{t}}\,,
    \label{eq:unitary-update}
\end{equation}
with $\eta$ the learning rate. This exponential map update guarantees $U_{t+1}$ remains unitary. In implementation, $\exp(\eta B)$ can be efficiently approximated using a truncated Taylor series or a scaling-and-squaring algorithm, and we re-normalize $U$ as needed to correct any numerical drift from unitarity.

\textbf{Fine-Tuning Algorithm:} Algorithm~\ref{alg:urnn-finetune} outlines the overall fine-tuning process using uRNN principles. We apply $U$ in the forward pass of the chosen transformer layer and then update $U$ using the above rule at each training step, while leaving other model weights to update as usual. Notably, although uRNNs were originally devised for recurrent sequence models, our strategy applies these unitary constraints to feed-forward or attention layers in a Transformer architecture. Conceptually, each forward pass through a transformer sub-layer is analogous to a single RNN step, with the sub-layer’s input playing the role of the “hidden state.” Maintaining $U$ as unitary thus helps preserve gradient norms through the depth of the network, even without explicit recurrence~\cite{bernardy2022assessing}.\footnote{In practice, we treat the input to each unitary-constrained sublayer as a proxy for an RNN hidden state, which ensures stable backpropagation across many transformer layers.}

The above integration of uRNN principles into Transformer fine-tuning offers several notable benefits. \textit{First}, enforcing unitary transformations provides \textbf{gradient stability}: it prevents the magnitudes of gradients from vanishing or exploding, even in very deep networks or tasks with long-range dependencies. 

\textit{Second}, this method tends to \textbf{improve convergence} during fine-tuning, as stable gradient norms facilitate faster and more reliable training (reducing the number of iterations required to reach a given performance level).

\textit{Third}, the approach is \textbf{adaptable} to different model components; unitary constraints can be applied to various layers or sub-layers of an LLM (e.g., attention projections or feed-forward blocks) without architecture changes, extending the use of orthogonal transformations beyond their traditional recurrent setting. By extending unitary transformations to transformer-based LLMs, we establish a novel paradigm for fine-tuning that marries parameter efficiency with training stability.

\subsection{Hybrid Fine-Tuning Approach}

We propose a per-layer fusion of the LoRA-GA and BOFT updates to capture both low-rank and orthonormal adaptation patterns. Specifically, this hybrid strategy computes the gradient-aligned low-rank update from LoRA-GA and the structured orthonormal update from BOFT for the same weight matrix in each layer, then mixes them with a dynamic coefficient. Intuitively, this allows fast initial adaptation via the low-rank component while gradually shifting emphasis to the BOFT component to stabilize learning as training proceeds.

\begin{algorithm}[htbp]
\caption{Hybrid Fine-Tuning Procedure \label{alg:hybrid_update}}
\small
\begin{algorithmic}[1]
    \STATE \textbf{Initialize:} pretrained weights $\{\mathbf{W}^\ell\}$, low-rank matrices $\{\mathbf{A}^\ell, \mathbf{B}^\ell\}$ for LoRA-GA, skew-symmetric matrices $\{\mathbf{Q}^\ell\}$ for BOFT.
    \STATE Set learning rates $\eta_{\text{LoRA}}, \eta_{\text{BOFT}}$; choose rank $r$, total epochs $E$.
    \FOR{$epoch = 1$ \TO $E$}
        \FOR{each minibatch in the dataset}
            \STATE \textbf{Forward Pass:}
            \FOR{each transformer layer $\ell$}
                \STATE Compute LoRA-GA update: \quad$\Delta \mathbf{W}_{\mathrm{LoRA}}^\ell = \mathbf{A}^\ell \mathbf{B}^\ell$.
                \STATE Compute BOFT orthonormal matrix:\\[0.5ex]
                    \quad$\mathbf{R}^\ell = (\mathbf{I}+\eta_{\text{BOFT}}\mathbf{Q}^\ell)(\mathbf{I}-\eta_{\text{BOFT}}\mathbf{Q}^\ell)^{-1}$.
                \STATE Compute BOFT update:\\[0.5ex]
                    \quad$\Delta \mathbf{W}_{\mathrm{BOFT}}^\ell = (\mathbf{R}^\ell - \mathbf{I})\mathbf{W}^\ell$.
            \ENDFOR
            \STATE Compute task-specific loss $\mathcal{L}$ using model predictions.
            \STATE \textbf{Backward Pass:}
            \FOR{each transformer layer $\ell$}
                \STATE Compute gradient norms:\\[0.5ex]
                    \quad $g_{\text{LoRA}}^\ell = \|\nabla_{\mathbf{A}^\ell,\mathbf{B}^\ell}\mathcal{L}\|$, \quad $g_{\text{BOFT}}^\ell = \|\nabla_{\mathbf{Q}^\ell}\mathcal{L}\|$.
                \STATE Compute dynamic weighting coefficient:\\[0.5ex]
                    \quad$\lambda^\ell = \frac{g_{\text{LoRA}}^\ell}{g_{\text{LoRA}}^\ell + g_{\text{BOFT}}^\ell}$.
                \STATE Form hybrid update for layer $\ell$:\\[0.5ex]
                    \quad$\Delta \mathbf{W}_{\mathrm{hybrid}}^\ell = \lambda^\ell \Delta \mathbf{W}_{\mathrm{LoRA}}^\ell + (1-\lambda^\ell)\Delta \mathbf{W}_{\mathrm{BOFT}}^\ell$.
                \STATE Update weight matrix for layer $\ell$:\\[0.5ex]
                    \quad$\mathbf{W}^\ell \leftarrow \mathbf{W}^\ell + \Delta \mathbf{W}_{\mathrm{hybrid}}^\ell$.
                \STATE Update low-rank matrices via gradient descent:\\[0.5ex]
                    \quad$\mathbf{A}^\ell \leftarrow \mathbf{A}^\ell - \eta_{\text{LoRA}}\nabla_{\mathbf{A}^\ell}\mathcal{L}$,\\[0.5ex]
                    \quad$\mathbf{B}^\ell \leftarrow \mathbf{B}^\ell - \eta_{\text{LoRA}}\nabla_{\mathbf{B}^\ell}\mathcal{L}$.
                \STATE Compute skew-symmetric gradient matrix for BOFT:\\[0.5ex]
                    \quad$\mathbf{G}^\ell = \nabla_{\mathbf{Q}^\ell}\mathcal{L} - (\nabla_{\mathbf{Q}^\ell}\mathcal{L})^\top$.
                \STATE Update skew-symmetric matrix $\mathbf{Q}^\ell$:\\[0.5ex]
                    \quad$\mathbf{Q}^\ell \leftarrow \mathbf{Q}^\ell - \eta_{\text{BOFT}}\mathbf{G}^\ell$.
                \STATE Recompute orthonormal matrix $\mathbf{R}^\ell$ via Cayley transform (for numerical stability):\\[0.5ex]
                    \quad$\mathbf{R}^\ell \leftarrow (\mathbf{I}+\eta_{\text{BOFT}}\mathbf{Q}^\ell)(\mathbf{I}-\eta_{\text{BOFT}}\mathbf{Q}^\ell)^{-1}$.
            \ENDFOR
            \STATE Update other model parameters (if any) via standard gradient descent.
        \ENDFOR
    \ENDFOR
\end{algorithmic}
\end{algorithm}

\textbf{Mathematical formulation:} 
Consider a layer $\ell$ with pretrained weight matrix $\mathbf{W}^\ell\in\mathbb{R}^{d_{\text{out}}\times d_{\text{in}}}$. We introduce low-rank factors $\mathbf{A}^\ell\in\mathbb{R}^{d_{\text{out}}\times r}$ and $\mathbf{B}^\ell\in\mathbb{R}^{r\times d_{\text{in}}}$ (rank $r$) as in LoRA~\cite{wang2023lora}, and a skew-symmetric matrix $\mathbf{Q}^\ell\in\mathbb{R}^{d_{\text{out}}\times d_{\text{out}}}$ ($\mathbf{Q}^\ell=-{\mathbf{Q}^\ell}^\top$) as in BOFT~\cite{liu2024boft}. The low-rank LoRA-GA update is
\[
\Delta \mathbf{W}^\ell_{\mathrm{LoRA}} \;=\; \mathbf{A}^\ell \mathbf{B}^\ell.
\]
The orthonormal BOFT update is obtained via the Cayley transform~\cite{dao2019learning}:
\[
\mathbf{R}^\ell \;=\; (\mathbf{I} + \eta \mathbf{Q}^\ell)(\mathbf{I} - \eta \mathbf{Q}^\ell)^{-1}, \qquad \mathbf{Q}^\ell = -{\mathbf{Q}^\ell}^\top,
\]
which ensures $\mathbf{R}^\ell$ is orthonormal (for small step size $\eta$). The BOFT update to $\mathbf{W}^\ell$ is then
\[
\Delta \mathbf{W}^\ell_{\mathrm{BOFT}} \;=\; (\mathbf{R}^\ell - \mathbf{I}_{d_{\text{out}}})\,\mathbf{W}^\ell.
\]
We combine these with a layerwise mixing coefficient $\lambda_t^\ell\in[0,1]$ that adapts over training steps $t$. Specifically, we set
\[
\lambda_t^\ell \;=\; \frac{\|\nabla_{\mathbf{A}^\ell,\mathbf{B}^\ell}L(\theta_t)\|}{\|\nabla_{\mathbf{A}^\ell,\mathbf{B}^\ell}L(\theta_t)\| + \|\nabla_{\mathbf{Q}^\ell}L(\theta_t)\|},
\]
so that the component with larger gradient norm receives higher weight. The hybrid update is then
\[
\Delta \mathbf{W}^\ell_{\mathrm{hybrid}} \;=\; \lambda_t^\ell\,\Delta \mathbf{W}^\ell_{\mathrm{LoRA}} + (1-\lambda_t^\ell)\,\Delta \mathbf{W}^\ell_{\mathrm{BOFT}},
\]
and the weight is updated as $\mathbf{W}^\ell \leftarrow \mathbf{W}^\ell + \Delta \mathbf{W}^\ell_{\mathrm{hybrid}}$. Here $\nabla_{\mathbf{A}^\ell,\mathbf{B}^\ell}L$ denotes the gradient of the training loss $L(\theta_t)$ with respect to the LoRA parameters~\cite{pfeiffer2020adapterfusion} $(\mathbf{A}^\ell,\mathbf{B}^\ell)$ and $\nabla_{\mathbf{Q}^\ell}L$ the gradient with respect to $\mathbf{Q}^\ell$. All notation above is defined per layer $\ell$.

\textbf{Pseudocode Algorithm:} Algorithm~\ref{alg:hybrid_update} summarizes the hybrid fine-tuning update. At each iteration, we compute both the LoRA-GA and BOFT updates for each layer, compute the mixing coefficient $\lambda_t^\ell$, and apply the weighted combination to update the weights.

The per-layer hybrid fusion adds only modest overhead beyond the individual LoRA-GA and BOFT updates. In each layer, the low-rank update costs $\mathcal{O}(d_{\text{out}}r + r\,d_{\text{in}})$ and the BOFT transform costs $\mathcal{O}(d_{\text{out}}\log d_{\text{out}})$. Computing $\lambda_t^\ell$ requires only the norms of gradients already computed, which is negligible. 

The total number of tunable parameters is the sum of the LoRA factors and any BOFT parameters (e.g., butterfly factors), comparable to using the two methods independently. By construction the Cayley parameterization enforces $\mathbf{R}^\ell$ to be orthonormal, which helps preserve gradient norms during optimization. The hybrid update thus integrates fast low-rank adaptation with structured orthogonal adjustments in a unified step.

%% file: Experiments.tex
This section evaluates parameter-efficient fine-tuning strategies in both general-purpose and multilingual settings. The analysis focuses on two aspects: effectiveness under limited supervision and deployability under realistic compute budgets. We report accuracy, calibration, and cross-language parity together with training footprint, training dynamics, and the relationship between cost and quality. Unless stated otherwise, each number is the mean of three runs with fixed seeds and identical optimization settings across methods.

\subsection{Setup and Evaluation Protocol}
\label{subsec:setup}

\textbf{Models.} We consider four open models that span size and multilingual coverage: Llama3.1-405B (405B parameters, long-context transformer)~\cite{meta2024llama3}, Llama3.3-70B (70B)~\cite{meta2024llama3}, Wizard-Vicuna-30B (30B, multilingual)~\cite{xu2023wizardlm,vicuna2023}, and BloomZ-7B1 (7.1B, multilingual)~\cite{muennighoff2022bloomz}.

\textbf{Methods.} The comparison includes Full Fine-Tuning (Full FT), LoRA, BOFT, LoRA with gradient alignment (LoRA-GA), unitary constraints inspired by uRNN (uRNN), and the proposed Hybrid method that combines low-rank gradient alignment with structured orthogonal updates. Hyperparameters follow a single protocol across tasks: LoRA rank $r{=}16$ with scaling $\alpha{=}32$; BOFT uses butterfly depth $m{=}3$; Hybrid applies a dynamic gradient weight $\alpha_t$.

\textbf{Benchmarks and metrics.} We use GLUE (macro accuracy across MNLI, QQP, SST-2, and QNLI)~\cite{wang2019glue}, GSM8K (math reasoning accuracy)~\cite{cobbe2021gsm8k}, MT-Bench (BLEU used as a multilingual translation proxy)~\cite{zheng2023mtbench}, and HumanEval for code generation measured by \emph{pass@1}~\cite{chen2021humaneval}. For multilingual low-resource evaluation we report XNLI (accuracy on English, Chinese, and Hindi)~\cite{conneau2018xnli} and FLORES (detokenized BLEU for EN$\to$ZH and EN$\to$ES)~\cite{goyal2021flores}. Cross-language parity is summarized by the Parity Gap $\Delta$, defined as the maximum minus the minimum score across languages, where smaller is better. Calibration is measured by the average Expected Calibration Error (Avg-ECE, in percentage), where smaller is better. 

\textbf{Compute.} Experiments run on nodes with dual AMD EPYC 7742 CPUs, 1~TB RAM, and 8$\times$NVIDIA A100 GPUs connected by NVLink and InfiniBand. The software stack uses PyTorch~2.0, CUDA~11.8, and NVIDIA Apex for mixed-precision training.

\subsection{General-Purpose Results}
\label{subsec:main_results}

Table~\ref{tab:all_results} compares all methods on GLUE, GSM8K, MT-Bench, and HumanEval. Hybrid tracks Full FT closely while keeping the number of trainable parameters small. On GLUE, Hybrid averages 92.3\% on the 405B model, which is 0.2 points below Full FT and 1.0 point above LoRA. On GSM8K, Hybrid reaches 55.9\% on 405B, slightly higher than Full FT and 1.7 points above LoRA. MT-Bench shows consistent BLEU gains over LoRA, BOFT, and LoRA-GA across sizes; on the 70B model Hybrid achieves 28.1 BLEU. On HumanEval, Hybrid attains 62.5\% \emph{pass@1} on 405B and 40.5\% on BloomZ-7B1, narrowing the gap to Full FT while remaining parameter-efficient. These findings suggest that combining gradient alignment with structural constraints yields robust generalization without full-model updates.

Trends are stable across scale. Relative to LoRA, the Hybrid advantage on GLUE is 1.0 at 405B, 1.1 at 70B, 0.9 at 30B, and 1.1 at 7B1. On GSM8K the margins are 1.7, 1.5, 1.3, and 1.3 points. MT-Bench follows the same pattern; for example, at 70B the averages are 27.8 for Hybrid and 27.0 for LoRA. On HumanEval, the edge is consistent and moderate; at 405B the averages are 62.5 for Hybrid and 61.0 for LoRA, and at 70B the averages are 58.5 and 56.9. BOFT and LoRA-GA reduce part of the gap relative to LoRA on several tasks, but Hybrid remains the most uniform alternative to Full FT under the shared protocol.

\subsection{Multilingual and Low-resource Evaluation}
\label{subsec:multilingual_low_resource}

We evaluate few-shot multilingual adaptation with 32 labeled examples per language. For XNLI~\cite{conneau2018xnli}, we report per-language accuracy on English (EN), Chinese (ZH), and Hindi (HI), as well as Macro, Parity Gap $\Delta$ (max--min across languages; lower is better), and Avg-ECE. For FLORES~\cite{goyal2021flores}, we report detokenized BLEU for EN$\to$ZH and EN$\to$ES, their average, and the direction gap $\Delta$.

Table~\ref{tab:xnli_combined} shows consistent gains for \emph{Hybrid} across both backbones. Relative to LoRA, macro accuracy increases by $+0.8$ to $+1.6$ points; relative to BOFT, by $+0.6$ to $+1.0$. Improvements are not concentrated in a single language: gains are largest on HI (e.g., $+1.7$ over LoRA on BloomZ and $+1.7$ on Wizard-Vicuna), moderate on ZH, and smaller but steady on EN. This pattern suggests that the method transfers better to lower-resource or morphologically distinct conditions without sacrificing high-resource performance.

% --- TABLE I ---
\begin{table*}[p]
\centering
\caption{\textbf{Performance across GLUE, GSM8K, MT-Bench, and HumanEval (\textit{pass@1)}.}
GLUE/GSM8K/MT-Bench report three runs and the average (Avg). HumanEval cells list three \textit{pass@1} runs and Avg.}
\label{tab:all_results}

\begingroup
\renewcommand{\arraystretch}{0.92}          
\setlength{\aboverulesep}{0.2ex}           
\setlength{\belowrulesep}{0.2ex}
\setlength{\tabcolsep}{4.2pt}              

\newcommand{\secrow}[1]{\multicolumn{6}{@{}l}{\textbf{#1}}\\[-0.35ex]\midrule}

\scriptsize
\begin{tabular*}{\textwidth}{@{\extracolsep{\fill}} l l c c c c @{}}
\toprule
\textbf{Benchmark} & \textbf{Method} &
\makecell{\textbf{Llama3.1}\\\textbf{405B}} &
\makecell{\textbf{Llama3.3}\\\textbf{70B}} &
\makecell{\textbf{Wizard-}\\\textbf{Vicuna-30B}} &
\makecell{\textbf{BloomZ}\\\textbf{7B1}} \\
\midrule
\secrow{GLUE (\%)}
 & Full FT & \runsavg{91.0}{94.0}{92.5}{92.5} & \runsavg{90.0}{91.0}{91.1}{90.7} & \runsavg{87.8}{90.0}{89.0}{88.9} & \runsavg{84.9}{86.2}{86.3}{85.8} \\
 & LoRA    & \runsavg{90.2}{91.5}{92.2}{91.3} & \runsavg{88.9}{89.2}{89.3}{89.1} & \runsavg{87.3}{87.4}{87.8}{87.5} & \runsavg{83.7}{84.2}{84.1}{84.0} \\
 & BOFT    & \runsavg{91.5}{92.0}{91.6}{91.7} & \runsavg{89.1}{89.8}{89.3}{89.4} & \runsavg{87.7}{87.9}{87.8}{87.8} & \runsavg{84.0}{84.5}{84.4}{84.3} \\
 & LoRA-GA & \runsavg{91.4}{92.3}{92.0}{91.9} & \runsavg{89.4}{89.8}{89.6}{89.6} & \runsavg{87.6}{87.9}{88.2}{87.9} & \runsavg{84.1}{84.3}{84.8}{84.4} \\
 & uRNN    & \runsavg{90.1}{91.5}{91.1}{90.9} & \runsavg{88.0}{88.8}{88.7}{88.5} & \runsavg{86.0}{86.7}{87.5}{86.7} & \runsavg{82.6}{83.9}{84.0}{83.5} \\
 & Hybrid  & \runsavg{91.7}{93.1}{92.1}{92.3} & \runsavg{89.8}{90.3}{90.4}{90.2} & \runsavg{87.9}{88.6}{88.7}{88.4} & \runsavg{84.6}{85.2}{85.4}{85.1} \\
\secrow{GSM8K (\%)}
 & Full FT & \runsavg{55.6}{56.0}{55.5}{55.7} & \runsavg{53.0}{53.5}{52.8}{53.1} & \runsavg{51.3}{51.9}{51.2}{51.5} & \runsavg{48.7}{49.0}{49.0}{48.9} \\
 & LoRA    & \runsavg{53.8}{54.4}{54.3}{54.2} & \runsavg{51.1}{51.9}{51.5}{51.5} & \runsavg{50.6}{51.0}{50.8}{50.8} & \runsavg{47.8}{48.2}{48.0}{48.0} \\
 & BOFT    & \runsavg{54.7}{55.0}{54.7}{54.8} & \runsavg{51.9}{52.1}{52.0}{52.0} & \runsavg{51.0}{51.2}{51.4}{51.2} & \runsavg{48.4}{48.5}{48.6}{48.5} \\
 & LoRA-GA & \runsavg{54.2}{54.8}{54.8}{54.6} & \runsavg{52.1}{52.4}{52.0}{52.2} & \runsavg{51.3}{51.6}{51.2}{51.4} & \runsavg{48.5}{48.8}{48.7}{48.7} \\
 & uRNN    & \runsavg{54.2}{54.7}{54.6}{54.5} & \runsavg{51.7}{51.9}{51.8}{51.8} & \runsavg{50.7}{51.1}{51.2}{51.0} & \runsavg{48.0}{48.4}{48.2}{48.2} \\
 & Hybrid  & \runsavg{55.3}{56.0}{56.4}{55.9} & \runsavg{52.8}{53.1}{53.0}{53.0} & \runsavg{51.7}{52.1}{52.5}{52.1} & \runsavg{48.7}{49.4}{49.8}{49.3} \\
\secrow{MT-Bench (BLEU)}
 & Full FT & \runsavg{28.7}{29.8}{29.4}{29.3} & \runsavg{27.5}{28.2}{27.9}{27.9} & \runsavg{26.0}{26.8}{26.4}{26.4} & \runsavg{24.5}{25.0}{24.8}{24.8} \\
 & LoRA    & \runsavg{28.4}{29.1}{28.5}{28.7} & \runsavg{27.0}{27.6}{27.3}{27.3} & \runsavg{25.4}{26.0}{25.9}{25.8} & \runsavg{23.8}{24.4}{24.7}{24.3} \\
 & BOFT    & \runsavg{28.6}{29.2}{29.0}{28.9} & \runsavg{27.1}{27.8}{27.5}{27.5} & \runsavg{25.6}{26.2}{26.0}{26.0} & \runsavg{24.0}{24.6}{24.9}{24.5} \\
 & LoRA-GA & \runsavg{28.7}{29.3}{29.1}{29.0} & \runsavg{27.2}{27.8}{27.7}{27.7} & \runsavg{25.9}{26.4}{26.2}{26.2} & \runsavg{24.2}{24.7}{25.2}{24.7} \\
 & uRNN    & \runsavg{28.2}{29.1}{28.5}{28.6} & \runsavg{26.7}{27.5}{27.1}{27.1} & \runsavg{25.2}{26.1}{25.7}{25.7} & \runsavg{23.7}{24.5}{24.4}{24.2} \\
 & Hybrid  & \runsavg{29.0}{29.6}{29.7}{29.4} & \runsavg{27.8}{28.3}{28.1}{28.1} & \runsavg{26.4}{26.8}{27.0}{26.7} & \runsavg{25.1}{25.7}{25.4}{25.4} \\
\secrow{HumanEval (\textit{pass@1}, \%)}
 & Full FT & \runsavg{61.8}{62.2}{62.0}{62.0} & \runsavg{\textbf{57.9}}{\textbf{58.8}}{\textbf{58.6}}{58.4} & \runsavg{54.0}{54.3}{54.1}{54.1} & \runsavg{41.1}{41.5}{41.3}{41.3} \\
 & LoRA    & \runsavg{60.8}{61.2}{61.0}{61.0} & \runsavg{\textbf{56.4}}{\textbf{57.3}}{\textbf{56.9}}{56.9} & \runsavg{51.9}{52.2}{52.0}{52.0} & \runsavg{39.0}{39.7}{39.5}{39.5} \\
 & BOFT    & \runsavg{61.3}{61.6}{61.4}{61.4} & \runsavg{\textbf{56.8}}{\textbf{57.6}}{\textbf{57.2}}{57.2} & \runsavg{52.3}{52.7}{52.5}{52.5} & \runsavg{39.6}{40.1}{39.9}{39.9} \\
 & LoRA-GA & \runsavg{61.4}{61.7}{61.5}{61.5} & \runsavg{\textbf{57.1}}{\textbf{57.9}}{\textbf{57.5}}{57.5} & \runsavg{52.6}{52.8}{52.7}{52.7} & \runsavg{39.5}{39.9}{39.7}{39.7} \\
 & uRNN    & \runsavg{60.5}{61.1}{60.8}{60.8} & \runsavg{\textbf{55.8}}{\textbf{56.7}}{\textbf{56.3}}{56.3} & \runsavg{51.6}{52.2}{51.9}{51.9} & \runsavg{38.6}{39.4}{39.0}{39.0} \\
 & Hybrid  & \runsavg{62.1}{62.9}{62.5}{62.5} & \runsavg{\textbf{58.0}}{\textbf{58.9}}{\textbf{58.6}}{58.5} & \runsavg{53.2}{53.8}{53.5}{53.5} & \runsavg{40.2}{40.8}{40.5}{40.5} \\
\bottomrule
\end{tabular*}
\endgroup
\end{table*}

% --- TABLE II: XNLI combined (BloomZ-7B1 + Wizard-30B) ---
\begin{table*}[!htbp]
\centering
\caption{XNLI, 32-shot per language. Accuracy (\%), Macro, Parity Gap $\Delta$ (\%), Avg-ECE (\%).}
\label{tab:xnli_combined}
\footnotesize
\begin{tabular}{l
S[table-format=2.1]S[table-format=2.1]S[table-format=2.1]S[table-format=2.1]S[table-format=1.1]S[table-format=1.1]
S[table-format=2.1]S[table-format=2.1]S[table-format=2.1]S[table-format=2.1]S[table-format=1.1]S[table-format=1.1]}
\toprule
 & \multicolumn{6}{c}{\textbf{BloomZ-7B1}} & \multicolumn{6}{c}{\textbf{Wizard-Vicuna-30B}}\\
\cmidrule(lr){2-7}\cmidrule(lr){8-13}
\textbf{Method} & {EN} & {ZH} & {HI} & {Macro} & {$\Delta$} & {Avg-ECE} & {EN} & {ZH} & {HI} & {Macro} & {$\Delta$} & {Avg-ECE}\\
\midrule
LoRA    & 80.6 & 77.9 & 73.5 & 77.3 & 7.1 & 2.8 & 84.7 & 81.9 & 77.2 & 81.3 & 7.5 & 2.2 \\
BOFT    & 81.2 & 78.4 & 74.1 & 77.9 & 7.1 & 2.5 & 85.1 & 82.4 & 77.8 & 81.8 & 7.3 & 2.0 \\
LoRA-GA & 81.4 & 78.6 & 74.3 & 78.1 & 7.1 & 2.4 & 85.4 & 82.7 & 78.0 & 82.0 & 7.4 & 1.9 \\
Hybrid  & 82.1 & 79.3 & 75.2 & 78.9 & 6.9 & 2.1 & 86.0 & 83.4 & 78.9 & 82.8 & 7.1 & 1.7 \\
\bottomrule
\end{tabular}
\end{table*}
\FloatBarrier

Cross-language balance slightly improves. On BloomZ-7B1, the Parity Gap narrows from $7.1$ to $6.9$; on Wizard-Vicuna-30B, from $7.5$ to $7.1$. Calibration also improves: Avg-ECE drops from $2.8\%$ to $2.1\%$ on BloomZ and from $2.2\%$ to $1.7\%$ on Wizard-Vicuna, indicating that probability estimates become more reliable while accuracy rises.

Table~\ref{tab:flores_combined} shows that translation trends mirror classification. \emph{Hybrid} yields average BLEU gains of about $+0.6$ over LoRA on BloomZ and about $+0.6$ over LoRA-GA on Wizard-Vicuna, while the direction gap $\Delta$ remains essentially unchanged (e.g., $7.7$ on BloomZ) or slightly shrinks (from $8.4$ to $8.2$ on Wizard-Vicuna). Quality gains therefore do not come at the expense of directional balance. Taken together, the results indicate that under the same 32-shot budget, \emph{Hybrid} improves accuracy and calibration uniformly across languages and tasks.

% --- TABLE III: FLORES combined (BloomZ-7B1 + Wizard-30B) ---
\begin{table*}[!htbp]
\centering
\caption{FLORES, 32-shot per language. Detokenized BLEU (\%), average, and direction gap $\Delta$ (\%).}
\label{tab:flores_combined}
\footnotesize
\begin{tabular}{l
S[table-format=2.1]S[table-format=2.1]S[table-format=2.1]S[table-format=1.1]
S[table-format=2.1]S[table-format=2.1]S[table-format=2.1]S[table-format=1.1]}
\toprule
 & \multicolumn{4}{c}{\textbf{BloomZ-7B1}} & \multicolumn{4}{c}{\textbf{Wizard-Vicuna-30B}}\\
\cmidrule(lr){2-5}\cmidrule(lr){6-9}
\textbf{Method} & {EN$\to$ZH} & {EN$\to$ES} & {Avg} & {$\Delta$} & {EN$\to$ZH} & {EN$\to$ES} & {Avg} & {$\Delta$} \\
\midrule
LoRA    & 24.6 & 32.3 & 28.5 & 7.7 & 27.4 & 35.8 & 31.6 & 8.4 \\
BOFT    & 25.1 & 32.8 & 29.0 & 7.7 & 27.9 & 36.2 & 32.0 & 8.3 \\
LoRA-GA & 25.3 & 33.0 & 29.2 & 7.7 & 28.1 & 36.4 & 32.2 & 8.3 \\
Hybrid  & 25.9 & 33.6 & 29.8 & 7.7 & 28.7 & 36.9 & 32.8 & 8.2 \\
\bottomrule
\end{tabular}
\end{table*}
\FloatBarrier

\subsection{Robustness to Orthographic Variants and Data Curation Effects}
\label{subsec:robustness_curation}

We evaluate two practical axes on XNLI. First, robustness to lightweight orthographic variants where each example is perturbed by exactly one operation (punctuation normalization, diacritics removal, or whitespace compaction). The metric is the macro-accuracy drop relative to clean input (lower is better). Second, governance-oriented data curation for BloomZ-7B1 under a 32-shot setting with three stages: C0 (raw crawl), C1 (+language identification and near-duplicate removal), and C2 (+light perplexity- and length-based filtering). We report macro accuracy, Parity Gap $\Delta$ (max–min across languages; lower is better), and Avg-ECE.

% --- TABLE: Orthographic noise robustness (combined) ---
\begin{table}[!htbp]
\centering
\caption{XNLI robustness to lightweight orthographic variants (macro accuracy drop, $\downarrow$~\%). Lower is better.}
\label{tab:xnli_robust_combined}
\footnotesize
\begin{tabular}{l S[table-format=1.1] S[table-format=1.1]}
\toprule
\textbf{Method} & {BloomZ-7B1} & {Wizard-30B} \\
\midrule
LoRA    & 3.2 & 2.7 \\
BOFT    & 2.6 & 2.3 \\
LoRA-GA & 2.5 & 2.2 \\
Hybrid  & 2.1 & 1.9 \\
\bottomrule
\end{tabular}
\end{table}
\FloatBarrier

Across both backbones, Hybrid shows the smallest degradation (2.1/1.9), improving over LoRA by $1.1/0.8$ points and remaining ahead of BOFT and LoRA-GA. The consistent ordering indicates higher invariance to frequent, low-severity input variations.

Curation produces monotonic gains: macro accuracy rises by $+1.5$ points from C0 to C2, while $\Delta$ and Avg-ECE drop by $0.6$ and $0.8$, respectively. Together with Table~\ref{tab:xnli_robust_combined}, these results suggest that the hybrid method reduces sensitivity to orthographic variation and that light, label-free governance improves cross-language parity and calibration under the same adaptation budget.

% --- TABLE: Data curation ablation ---
\begin{table}[!htbp]
\centering
\caption{Data curation ablation on XNLI (BloomZ-7B1, 32-shot).}
\label{tab:curation_ablation}
\footnotesize
\begin{tabular}{l S[table-format=2.1] S[table-format=1.1] S[table-format=1.1]}
\toprule
\textbf{Stage} & {Macro (\%) $\uparrow$} & {$\Delta$ (\%) $\downarrow$} & {Avg-ECE (\%) $\downarrow$} \\
\midrule
C0: Raw                 & 77.4 & 7.5 & 2.9 \\
C1: +LID +Dedup         & 78.2 & 7.2 & 2.5 \\
C2: +Quality Filtering  & 78.9 & 6.9 & 2.1 \\
\bottomrule
\end{tabular}
\end{table}
\FloatBarrier

%%%%%%%%%%%%%%%%%%%%%%%%%%%%%%%%%%%%%%%%%%%%%%%%%%%%%%%%

\subsection{Few-shot Scaling and Training Footprint}
\label{subsec:scaling_efficiency}

\begin{figure}[!htbp]
\centering
\includegraphics[width=1\columnwidth]{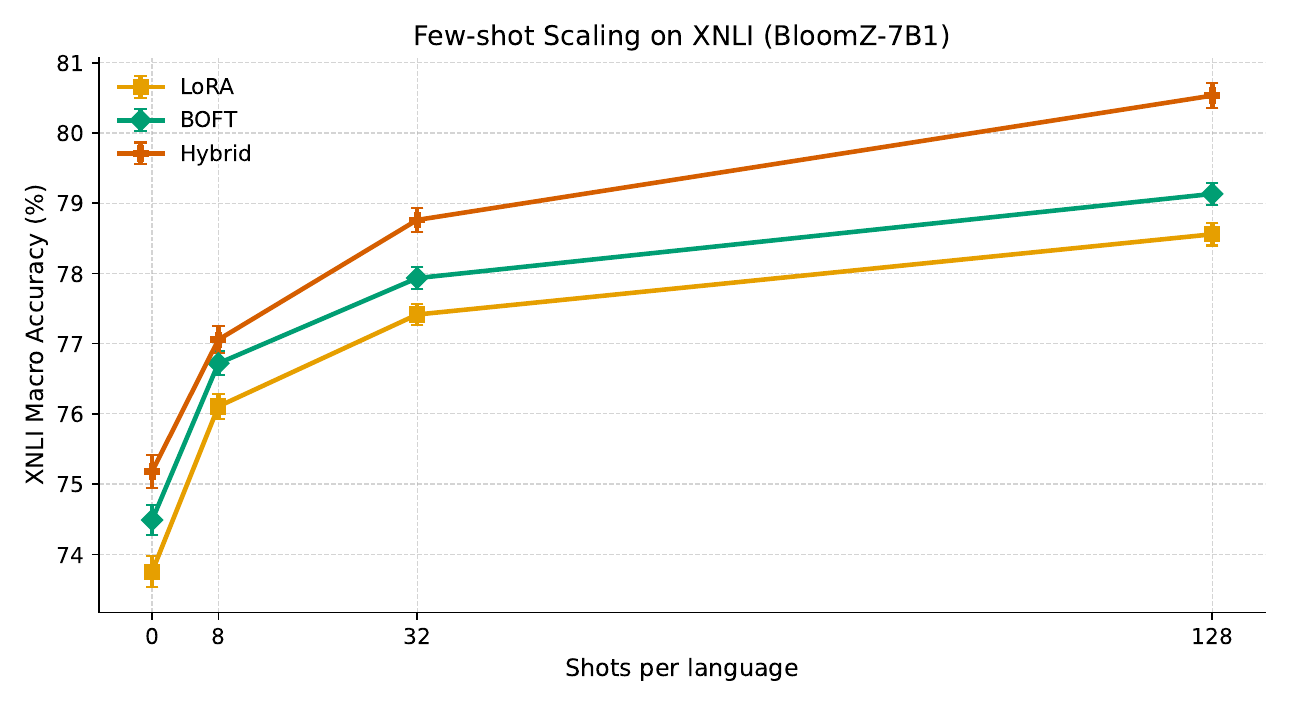}
\caption{Few-shot scaling on XNLI (BloomZ-7B1). Macro accuracy at 0, 8, 32, and 128 shots.}
\label{fig:fewshot_scaling}
\end{figure}

We study 0, 8, 32, and 128-shot scaling on XNLI with BloomZ-7B1 (Figure~\ref{fig:fewshot_scaling}). At zero-shot, \emph{Hybrid} has a small advantage; the margin widens to about $+1.6$ points at 32-shot and $+1.9$ points at 128-shot over LoRA.

\FloatBarrier

\begin{figure}[!htbp]
\centering
\includegraphics[width=1\columnwidth]{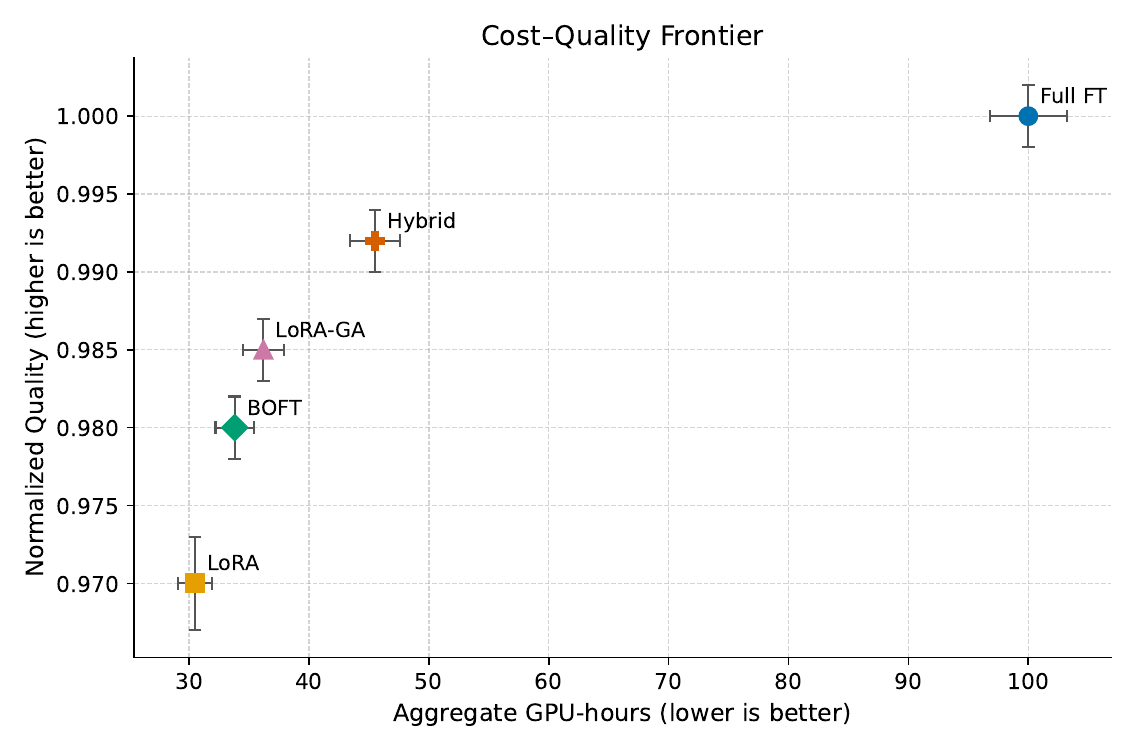}
\caption{Cost–quality frontier across methods. Upper-left is better.}
\label{fig:cost_quality_frontier}
\end{figure}

We then measure training footprint under the multilingual 32-shot protocol on Wizard-Vicuna-30B and BloomZ-7B1 (Table~\ref{tab:footprint_combined}) and summarize the cost–quality frontier (Figure~\ref{fig:cost_quality_frontier}). \emph{Hybrid} introduces modest overhead: approximately $0.6$–$0.7$~GB higher peak memory and $0.1$–$0.2$~s longer step time at comparable scales, with only tens of millions of additional tunable parameters. Given the accuracy and calibration gains, its points lie near the empirical Pareto front, indicating improved quality at nearly the same budget.

\subsection{Training Dynamics and Stability}
\label{subsec:dynamics}

\begin{figure}[!htbp]
\centering
\includegraphics[width=1\columnwidth]{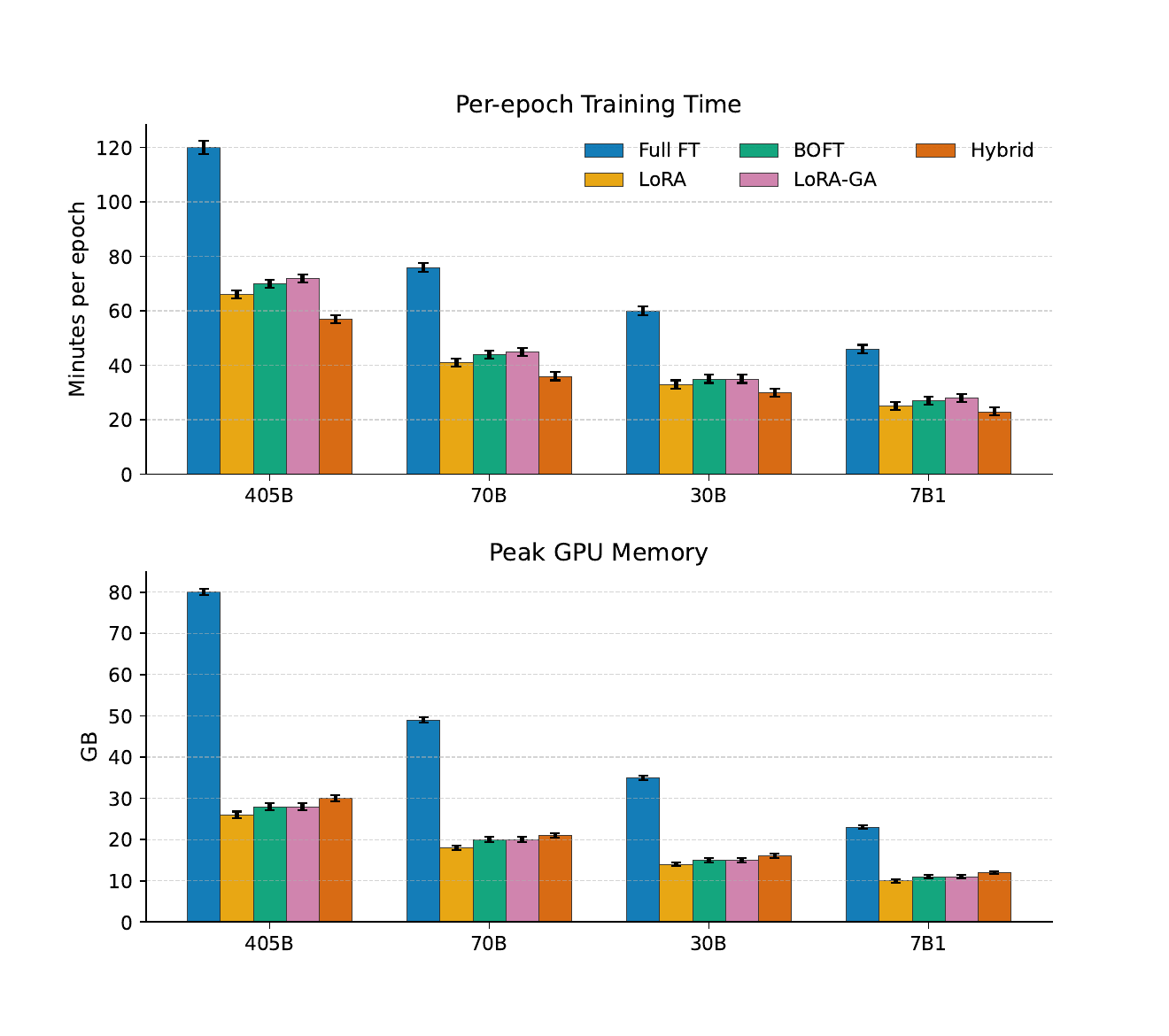}
\caption{Per-epoch training time and peak GPU memory per method across model scales.}
\label{fig:training_resource}
\end{figure}

Figure~\ref{fig:training_resource} reports per-epoch time and peak memory across scales, and Figure~\ref{fig:grad_stability} tracks gradient norms and validation loss over epochs on Wizard-Vicuna-30B. \emph{Hybrid} suppresses early spikes in gradient norm and descends faster in the first few epochs, narrowing the gap to Full FT by the mid-stage. Final validation loss reaches $0.88$ versus $0.83$ for Full FT, while using far fewer tunable parameters, indicating that most of the quality of full adaptation is recovered at a fraction of the cost.

% --- TABLE: Training footprint combined ---
\begin{table*}[!htbp]
\centering
\caption{Training footprint in the multilingual 32-shot setting. Peak memory (GB), step time (s), and tunable parameters (M).}
\label{tab:footprint_combined}
\footnotesize
\begin{tabular}{l
S[table-format=2.1]S[table-format=1.2]S[table-format=3.1]
S[table-format=2.1]S[table-format=1.2]S[table-format=3.1]}
\toprule
 & \multicolumn{3}{c}{\textbf{Wizard-Vicuna-30B}} & \multicolumn{3}{c}{\textbf{BloomZ-7B1}} \\
\cmidrule(lr){2-4}\cmidrule(lr){5-7}
\textbf{Method} & {Peak mem.} & {Step time} & {Tunable} & {Peak mem.} & {Step time} & {Tunable} \\
\midrule
LoRA    & 14.8 & 1.35 & 93.2  & 12.3 & 0.91 & 58.7 \\
BOFT    & 15.4 & 1.41 & 98.7  & 12.8 & 0.95 & 62.1 \\
LoRA-GA & 15.5 & 1.44 & 98.7  & 12.9 & 0.97 & 62.1 \\
Hybrid  & 16.1 & 1.51 & 104.3 & 13.4 & 1.02 & 66.0 \\
\bottomrule
\end{tabular}
\end{table*}

\FloatBarrier

LoRA and LoRA-GA start with larger gradients and exhibit a short plateau before resuming improvement, matching the smaller few-shot gains observed in the scaling study. BOFT and uRNN maintain tighter gradients but converge to slightly higher losses, consistent with their weaker macro accuracy. The fused design of \emph{Hybrid} combines rapid early progress with stable late-phase optimization, which aligns with its robustness and parity improvements: the method learns quickly enough to exploit limited supervision while avoiding destabilizing updates later in training.

%% file: Conclusion.tex
This work presents a \emph{governance-aware} recipe for multilingual, low-resource adaptation of Large Language Models through a \emph{Hybrid} Parameter-Efficient Fine-Tuning (PEFT) scheme with selective unitary stabilization. Rather than introducing yet another isolated adapter, we studied how a single-step, per-layer fusion behaves under realistic constraints in which accuracy must be accompanied by calibration, cross-language parity, robustness to routine input variants, and a budgeted training footprint. Across general-purpose tasks, \emph{Hybrid} tracks full-model fine-tuning while maintaining a small trainable set, and on code generation reaches 62.5\% \emph{pass@1} at 405B. In 32-shot multilingual adaptation, it improves XNLI macro accuracy by +0.8--1.6 over LoRA and +0.6--1.0 over BOFT, and increases FLORES average BLEU by +0.6 without widening direction gaps. It further exhibits the smallest degradation under lightweight orthographic variants (2.1/1.9~pp drops on BloomZ-7B1/Wizard-Vicuna-30B), and stays near a favorable cost--quality frontier with only modest overhead (+0.6--0.7~GB peak memory and +0.1--0.2~s step time). Together with reduced calibration error and narrower parity gaps under simple curation, these findings indicate that mixing gradient-aligned low-rank updates with structured orthogonal adjustments---and reinforcing deep stacks via selective unitary constraints---yields a stable, calibrated, and resource-efficient path for multilingual fine-tuning under tight budgets.

Beyond headline metrics, our analysis highlights two practical aspects. First, the fused update suppresses early gradient spikes and accelerates mid-stage descent while preserving late-epoch stability, which helps recover much of the full-tuning quality at a fraction of the trainable parameters. Second, the evaluation protocol---pairing accuracy with Expected Calibration Error, parity gaps, orthographic robustness, and light, label-free data stewardship---offers a compact but actionable lens for deployment readiness, particularly when practitioners operate under fixed memory and time budgets. The overall recipe remains simple: a per-layer, single-step combination that is compatible with standard PEFT pipelines and amenable to existing mixed-precision and scheduling stacks.

Our findings point to several directions. \emph{(i) Learning the mixer.} The current gradient-norm mixing is simple and effective; learning the per-layer schedule (e.g., via meta-gradients or small controllers) under a joint accuracy--calibration--parity objective could further improve the frontier. \emph{(ii) Hardware-aware integration.} Combining \emph{Hybrid} with quantization-aware training (e.g., low-bit activations) \cite{jacob2018quantization} and memory schedulers may reduce the residual overhead while preserving stability in deep stacks. \emph{(iii) Broader multilingual stress-tests.} Extending beyond EN/ZH/HI to more scripts and families, and jointly measuring fairness-adjacent disparity metrics alongside ECE, would clarify when parity gains persist at scale \cite{blasi2022systematic}. \emph{(iv) Robustness under real noise.} Beyond orthographic tweaks, evaluating resilience to tokenization drift \cite{xue2022byt5}, locale-specific normalization, and web-crawl artifacts can better align with deployment realities.

\begin{figure*}[!t]
\centering
\includegraphics[width=1\textwidth]{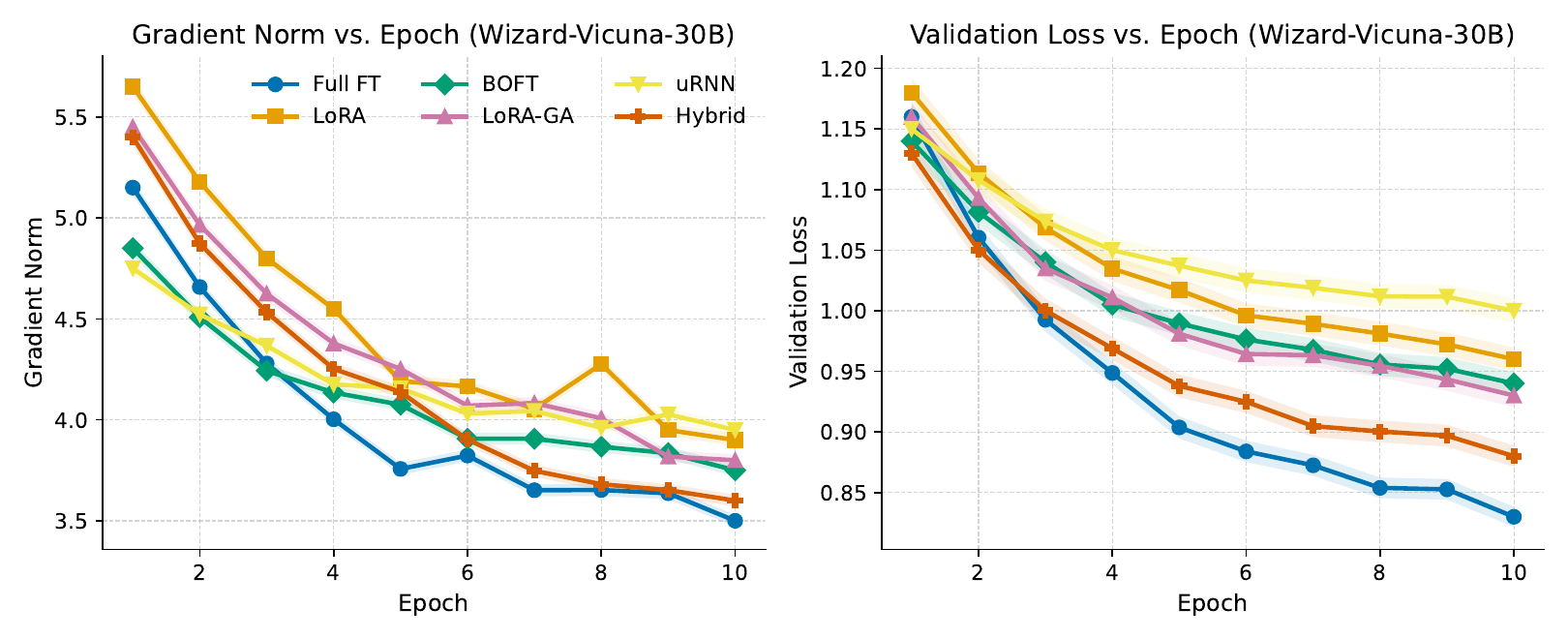}
\caption{Training stability on Wizard-Vicuna-30B across ten epochs: gradient norm and validation loss.}
\label{fig:grad_stability}
\end{figure*}
\FloatBarrier